\documentclass[numbers,preprint,review,12pt]{elsarticle}

\usepackage{hyperref}

\journal{Neurocomputing}

\usepackage{graphicx}
\usepackage{multirow}
\usepackage{times}
\usepackage{amsmath,amsthm}
\usepackage{amssymb}
\usepackage{amsfonts}
\usepackage{algorithm}
\usepackage{mathrsfs}
\usepackage{booktabs} 
\usepackage{color}
\usepackage{hyperref} 
\usepackage{url}
\usepackage{caption} 
\usepackage{algpseudocode}
\usepackage{pifont}
\usepackage{bm}
\usepackage{comment}
\usepackage{wasysym}
\usepackage{soul}
\usepackage{CJKutf8}

\begin{document}

\begin{frontmatter}



\title{RAGNav: A Retrieval-Augmented Topological Reasoning Framework for Multi-Goal Visual-Language Navigation}

\author[swufe]{Ling Luo}
\author[swufe]{Qianqian Bai}


\address[swufe]{Southwestern University of Finance and Economics, Chengdu, 611130, China}

\begin{abstract}





Vision-Language Navigation (VLN) is evolving from single-point pathfinding toward the more challenging Multi-Goal VLN. This task requires agents to accurately identify multiple entities while collaboratively reasoning over their spatial-physical constraints and sequential execution order. However, generic Retrieval-Augmented Generation (RAG) paradigms often suffer from spatial hallucinations and planning drift when handling multi-object associations due to the lack of explicit spatial modeling.To address these challenges, we propose RAGNav, a framework that bridges the gap between semantic reasoning and physical structure. The core of RAGNav is a Dual-Basis Memory system, which integrates a low-level topological map for maintaining physical connectivity with a high-level semantic forest for hierarchical environment abstraction. Building on this representation, the framework introduces an anchor-guided conditional retrieval and a topological neighbor score propagation mechanism. This approach facilitates the rapid screening of candidate targets and the elimination of semantic noise, while performing semantic calibration by leveraging the physical associations inherent in the topological neighborhood.This mechanism significantly enhances the capability of inter-target reachability reasoning and the efficiency of sequential planning. Experimental results demonstrate that RAGNav achieves state-of-the-art (SOTA) performance in complex multi-goal navigation tasks.

\end{abstract}


\end{frontmatter}


\section{Introduction}

Vision-and-Language Navigation (VLN) aims to enable an agent to move in complex environments by following natural language instructions, serving as a key task connecting perception, language, and action~\cite{anderson2018vision}. In recent years, the task paradigm has expanded from single-goal navigation to Multi-goal VLN, which is more practically valuable and challenging~\cite{wani2020multion,raychaudhuri2024mopa,sadek2023multi}. Such tasks (e.g., ``First go to the bedside in the bedroom, then to the desk in the study") require the agent not only to identify individual goals but also to perform joint reasoning and sequential planning on the spatial relationships and semantic relationships among multiple goals.

To conduct exploratory planning in complex environments, constructing a lightweight internal representation of the environment is crucial. Topological maps, which abstract continuous environments into graphs composed of key locations (nodes) and their connectivity (edges), have become a widely adopted form of spatial memory in VLN~\cite{gao2021room,giuliari2022spatial,zeng2023multi}. However, traditional topological maps mainly encode geometric or low-level visual features, exposing inherent limitations when dealing with multi-goal instructions. First, they are poor in semantic representation. Topological nodes are usually difficult to associate with high-level semantic concepts mentioned in instructions (e.g., ``living room'', ``desk'') because their construction process often ignores semantic category information critical for task reasoning~\cite{dang2022unbiased,liu2023revolt}.Second, they are weak in relational reasoning. Although spatial accessible paths can be found through graph search, existing methods struggle to understand the semantic context between goals, leading to semantically unreasonable paths or incorrect goal orders in planning. Essentially, traditional topological maps lack a queryable, multi-modal semantic memory system, making it impossible to effectively store and utilize rich visual, spatial, and semantic information from embodied experience.

Retrieval-Augmented Generation (RAG) technology significantly improves the factual accuracy and reasoning traceability of Large Language Models (LLMs) in knowledge-intensive tasks by equipping them with a retrievable external knowledge base~\cite{lewis2020retrieval,li2022survey}. This provides a new idea to bridge the semantic gap of topological maps: can we construct a semantic memory based on RAG to enhance or even partially replace the functions of traditional topological maps? Recent works such as EmbodiedRAG~\cite{booker2024embodiedrag} propose to enhance LLM-based planners by preprocessing task-related entities and attributes and retrieving associated information to execute natural language robot tasks. Such methods systematically introduce the ``retrieval-generation" mechanism of RAG into embodied scenarios for the first time, providing a feasible path for building queryable scene semantic memory.

However, directly applying RAG to multi-goal VLN still faces severe challenges. First, multimodal data querying poses a significant challenge. It still requires specialized design to efficiently organize heterogeneous data such as visual and spatial pose to enable accurate responses to complex navigation instruction queries~\cite{zhao2026retrieval,zhao2023retrieving}. Second, the challenge of spatial-semantic coupled retrieval. The vector similarity-based retrieval mechanism of NaiveRAG lacks explicit modeling of physical spatial layout and connectivity, making it unable to directly answer spatial relationship queries such as ``Is it accessible from A to B"~\cite{lewis2020retrieval}; while GraphRAG methods that can construct complex associations often fail to meet the real-time requirements of navigation due to their retrieval efficiency~\cite{edge2024local,peng2025graph}.

To address the above challenges, this paper proposes \textbf{RAGNav}, a retrieval-augmented topological semantic reasoning framework for multi-goal Visual-Language Navigation. The core innovation of this framework lies in the construction of a dual-basis environmental memory system jointly supported by a low-level topological map and a high-level semantic forest. In this system, the topological map serves as the physical skeleton of the environment, maintaining connectivity among key pose nodes and enforcing spatial topological constraints. Meanwhile, the semantic forest hierarchically indexes multimodal embodied experiences, enabling efficient organization and representation of environmental information at different granularities. Based on this dual-basis representation, we further propose a spatial--neighbor dual-dimensional retrieval-augmented strategy. In the spatial dimension, the system leverages the hierarchical structure of the semantic forest to perform rapid pruning, thereby achieving scale alignment and preliminary localization of target candidate regions within massive environmental data. In the neighbor dimension, the system deeply exploits physical correlations within topological neighborhoods, filters out isolated noise through a semantic fingerprint verification mechanism, and performs in-depth reasoning over the sequential reachability among multiple goals using graph propagation mechanisms. This strategy not only enables the RAG system to identify specific target attributes, but also allows it to deeply understand the complex spatial coupling logic among multiple goals. Ultimately, this paradigm drives a complete task loop composed of perception, planning, execution, and reflection, significantly enhancing the agent's robustness and decision-making efficiency when facing uncertain environments.

The main academic contributions of this paper can be summarized as follows
\begin{itemize}
    \item To address the mismatch between semantic logic and spatial topology in multi-goal VLN tasks, we propose a new model that leverages non-parametric memory to achieve hierarchical accumulation of environmental knowledge and logical reconstruction of long instructions.
    \item We propose the RAGNav framework, which effectively addresses the spatial-semantic gap by implementing hierarchical pruning in the semantic forest and reasoning across topological neighborhoods. This enables deep semantic alignment from high-level task parsing down to low-level physical verification.
    \item Experimental results demonstrate that the RAGNav framework achieves state-of-the-art performance on multi-goal navigation tasks. Its retrieval module significantly outperforms baseline methods such as NaiveRAG~\cite{lewis2020retrieval}, GraphRAG~\cite{edge2024local}, and LightRAG~\cite{guo2024lightrag} in both efficiency and accuracy.
\end{itemize}

\section{Related Work}
\subsection{Multi-Object Navigation}


Object goal navigation can be divided into single-object navigation~\cite{anderson2018vision,qi2020reverie,ku2020room} and Multi-Object Navigation~\cite{wani2020multion,raychaudhuri2024mopa,sadek2023multi} according to the number of task goals. As a generalization of the former, the latter requires the agent to explore and locate multiple targets in sequence in an unknown environment based on a given sequence of semantic labels. Its core challenge shifts from a single "localization-arrival" to the understanding of the spatial and semantic relationship network implied in the instructions, as well as the planning of the global action sequence. To address this challenge, existing studies have made certain breakthroughs. For example, Chen et al.~\cite{chen2022learning} first proposed an active camera strategy to solve the multi-object navigation problem, which significantly improved the success rate; Sadek et al.~\cite{sadek2023multi} proposed a modular hybrid navigation method that decouples high-level semantic planning from low-level motion control, focusing on the transfer of strategies from simulation to reality; Marza et al.~\cite{marza2023multi} recently proposed using a dual neural network to learn the neural implicit representation of scenes, aiming to dynamically construct and utilize the global contextual information of scenes, which provides a new paradigm for the multi-object goal navigation task.

However, the reasoning of these methods on the relationships among multiple targets is often offline or fragmented, failing to form a closed loop that can continuously and actively "retrieve" the next most relevant target or key path point from the environmental memory during the planning process. The lack of such a dynamic retrieval order limits the agent's reasoning ability in complex scenes and under lengthy instructions, and the planning is prone to fall into local optimality or semantic sequence errors. Therefore, there is an urgent need for a new framework that deeply couples spatial topological structure with semantic relationship retrieval to realize dynamic and interpretable reasoning on the access order of multiple targets. This work is precisely carried out to address this key issue.

\subsection{Retrieval Augmented
Generation}

Retrieval-Augmented Generation (RAG) ~\cite{lewis2020retrieval} is a core method integrating information retrieval and generation models. By introducing external knowledge sources and retrieval mechanisms, it effectively makes up for the limitations of traditional sequence-to-sequence frameworks and significantly improves the performance of natural language processing tasks. Since then, a variety of technologies and adaptive methods have been proposed, such as: MuRAG ~\cite{chen2022murag} which enhances question-answering capabilities by retrieving images, Self-RAG ~\cite{asai2024self} which introduces a self-reflection mechanism to optimize outputs, InstructRAG ~\cite{weiinstructrag} which adopts explicit denoising technology to filter noisy information, GRAFT-Net ~\cite{sun2018open} which fuses corpus texts and knowledge graphs to construct heterogeneous graphs, and ToG ~\cite{sunthink} which takes knowledge graphs as the index structure. However, the above studies mostly focus on pure text or general scenarios, while this paper focuses on the field of embodied navigation -- the robot can actively change the retrieval information source, and the information related to the original query will dynamically change as the robot executes the plan. In addition, the embodied navigation experience itself has the characteristics of redundancy, hierarchical correlation and spatial anchoring, which make the above pure text graph construction methods difficult to adapt and perform poorly.

\subsection{Navigation Environment Representation}

To achieve efficient planning in unknown environments, constructing a lightweight and semantically rich internal representation is the key. Traditional methods usually rely on static observations~\cite{zellers2018neural} and predefined semantic systems to build topological semantic maps~\cite{gao2021room,giuliari2022spatial,ost2021neural}, which makes it difficult for them to adapt to the fine-grained semantic requirements of open scenarios and lack the ability of online update~\cite{an2024etpnav}. Recent relevant studies have made remarkable progress. First, semantic injection based on Vision-Language Models (VLM) enables fast semantic annotation by attaching descriptive texts generated by VLM to map nodes~\cite{guo2025genesis}, but this method is only a loosely coupled semantic superposition, lacking explicit modeling of semantic relationships between objects, and thus hard to support tasks that require deep spatial-semantic joint reasoning. Second, the unified representation based on 3D scene graphs~\cite{zemskova20253dgraphllm} encodes objects, semantics and their relationships in a dense graph structure. Although it has strong reasoning potential, its construction relies on expensive 3D reconstruction technology, leading to low efficiency in online construction. In addition, the flat structure lacks hierarchical abstraction, which results in low retrieval efficiency when facing high-level semantic queries. Third, the exploration of hierarchical semantic maps~\cite{zhao2025slam} attempts to approach human cognition by constructing multi-layer structures such as "object-region" to improve planning efficiency. However, the high-level abstraction of existing methods mostly relies on fixed spatial segmentation or simple clustering rules, and their semantic division is static and rigid, which cannot evolve dynamically according to task contexts (e.g., aggregating scattered "coffee machines" and "sinks" into a "breakfast nook"). Meanwhile, they fail to solve the core problem of how to efficiently support dynamic retrieval and sequential reasoning of multi-target and multi-relation instructions based on this structure.

\section{Methodology}

The RAGNav framework proposed in this paper aims to address the challenges of long instruction decomposition and spatial relationship reasoning in multi-objective navigation. As illustrated in Figure~\ref{fig:RAGNav框架图}, the framework consists of two phases: offline environmental memory construction and an online task execution loop. In the offline phase, the framework utilizes pose data to construct environmental representations, forming a dual-baseline environmental memory composed of topological maps and semantic forests. During the online phase, the system employs a Large Language Model (LLM) to parse long instructions, decomposing them into subtask chains with logical dependencies while extracting the spatial and temporal dependencies involved. Leveraging a two-stage retrieval and topology enhancement mechanism, the system performs spatial consistency verification and precise associative localization for multiple target entities. By integrating feedback on topological connectivity, the system conducts sequential planning for multiple objectives, forming a closed-loop "perception-planning-execution-reflection" task cycle. This significantly enhances the execution efficiency of multi-objective tasks under complex instructions.

\begin{figure*}[htb]
  \centering
  \includegraphics[width=0.95\textwidth]{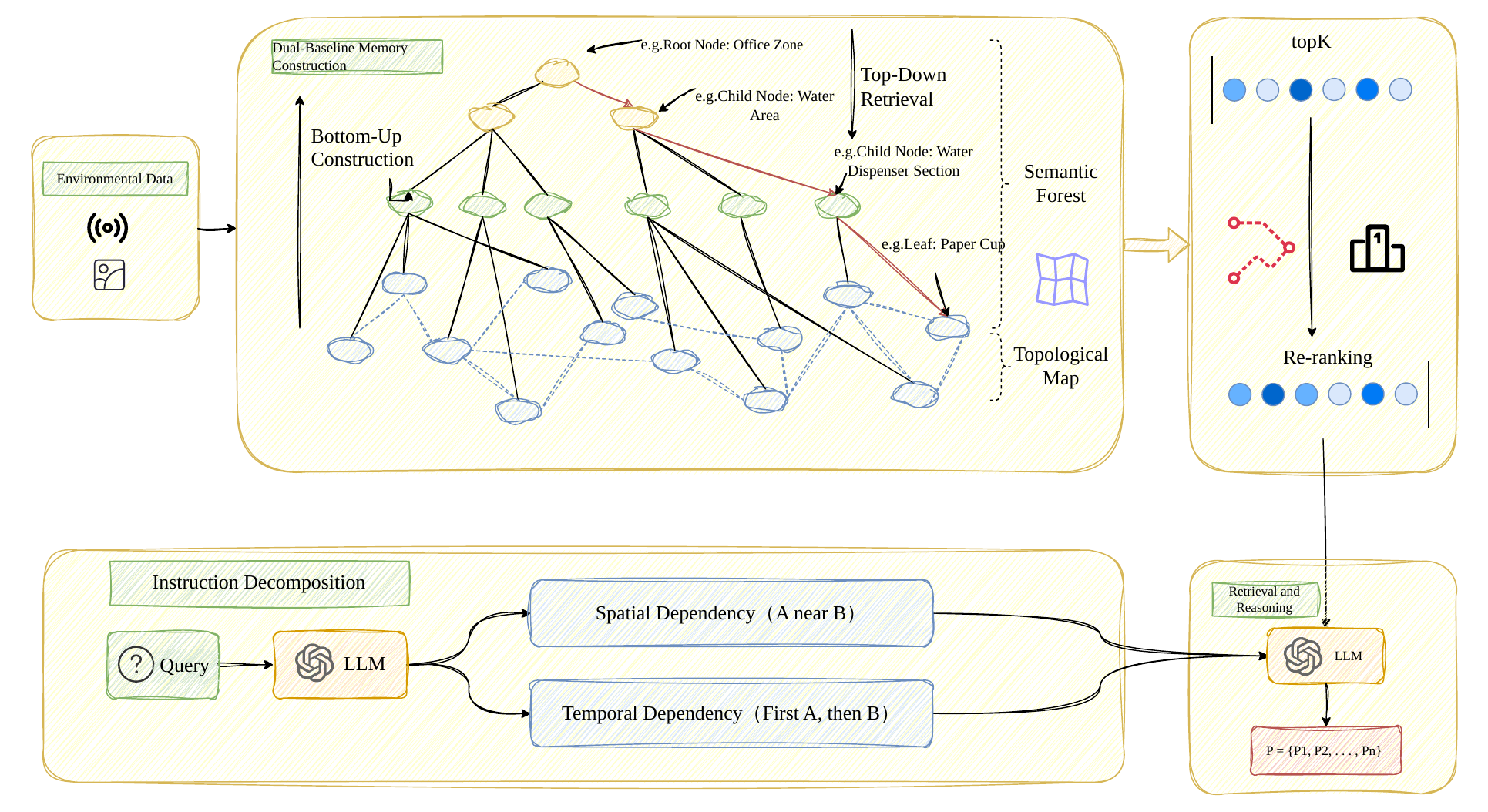}
  \vspace{8pt}
  \caption{Architecture of the RAGNav Framework}
  \label{fig:RAGNav_Architecture}
\end{figure*}

\subsection{Embodied Data Acquisition and Autonomous Exploration}

Prior to constructing the dual-basis environment memory, the framework establishes a comprehensive multimodal perception and autonomous mapping system. During exploration, the agent continuously collects synchronized data including timestamps $t$, RGB image observations $I_t$, LiDAR point clouds $P_t$, and six-degree-of-freedom (6-DoF) poses $\xi_t$ (comprising position $(x,y,z)$ and orientation) generated via odometry fusion. The precise geometric information provided by LiDAR point clouds is crucial for building the spatial skeleton of the environment and understanding free space. All raw data are transformed into a unified global world coordinate system, forming a spatiotemporally aligned raw data stream:
\begin{equation}
D = \{(t, I_t, P_t, \xi_t)\}
\end{equation}

To achieve efficient environment coverage, the system deploys an active mapping algorithm based on Frontier Exploration. The algorithm discretizes the continuous environment into grids with a resolution of 1.0 m and dynamically tracks exploration states. Based on the coverage of the current LiDAR frame $P_t$, grids are marked as “explored” or “unknown.” By scanning the four-connected neighborhood of explored grids, frontier candidate points located at the boundaries are identified. Breadth-first search (BFS) is then applied for spatial clustering to suppress noise, yielding stable frontier regions $C_i$. Each valid frontier region is represented by its geometric center as a frontier point $f_i$. The agent adopts a greedy strategy, selecting the frontier point closest to its current position $p_{\text{current}}$ as the next navigation target:
\begin{equation}
f_{\text{next}} = \arg\min_{f \in F} \|p_{\text{current}} - f\|_2
\end{equation}


\subsection{Intelligent Task Decomposition and Execution Planning}

To address the challenges of lengthy and logically coupled instructions in multi-goal navigation, the framework introduces an intelligent task decomposition and sequential execution mechanism. The system first utilizes a large language model (LLM) as the core reasoning module to parse the original natural language instruction $Q$ into a structured task sequence $\mathcal{T} = \{t_1, t_2, \dots, t_n\}$. During this process, dependency identification is performed, modeling inter-task constraints into two fundamental logical types that guide subsequent retrieval.

For spatial dependency, when instructions contain explicit spatial constraints (e.g., ``A near B''), the system designates target $B$ as the anchor and target $A$ as the condition-dependent entity. Retrieval then shifts from global blind search to a conditional probability maximization problem within the topological neighborhood $\mathcal{N}(v_B)$ of the anchor node $v_B$. The spatial score $S_{spatial}$ of a candidate node $v_i$ is defined as the product of semantic similarity and physical distance constraint:
\begin{equation}
S_{spatial}(v_i | v_B) = \text{sim}(\phi(Q_A), \psi(v_i)) \cdot \exp\left( -\frac{\|p_i - p_B\|^2}{2\sigma^2} \right)
\end{equation}
where $\phi(\cdot)$ and $\psi(\cdot)$ denote text and visual embedding functions, respectively, and $\text{sim}(\cdot)$ represents cosine similarity. The exponential term applies a Gaussian kernel to softly constrain physical distance, ensuring that candidate target $A$ lies spatially close to the anchor $B$ at position $p_B$, thereby filtering semantic noise.

For temporal dependency, when instructions specify an explicit visiting order (e.g., ``First A, then B''), the system formulates the problem as a cost-sensitive sequential planning task. Given the set of retrieved target locations $\mathcal{P} = \{p_1, p_2, \dots, p_n\}$, the objective is to find the globally optimal execution path $\Pi^*$:
\begin{equation}
\Pi^* = \arg\min_{\pi \in \text{Perm}(\mathcal{P})} \sum_{j=1}^{n-1} \mathcal{D}(p_{\pi(j)}, p_{\pi(j+1)}) + \lambda \cdot \mathcal{L}(\pi, \mathcal{S})
\end{equation}
Here, $\text{Perm}(\mathcal{P})$ denotes all permutations of candidate paths, $\mathcal{D}(\cdot)$ is the shortest-path cost computed from topological connectivity, and $\mathcal{L}(\cdot)$ is a semantic deviation penalty measuring inconsistency between execution order $\pi$ and the instruction-defined semantic sequence $\mathcal{S}$. The weight $\lambda$ balances physical efficiency and semantic alignment.

During execution, the system applies differentiated scheduling strategies based on identified dependencies. Independent tasks trigger parallel retrieval to minimize latency, while spatially dependent tasks strictly follow an “anchor-first, local-expansion” strategy. This hierarchical decomposition and sequential planning mechanism allows RAGNav to transform complex long-horizon tasks into a stream of spatiotemporally consistent subgoals, effectively addressing semantic drift and localization failure in multi-goal navigation.

\subsection{Task-Driven Dual-Basis Memory Construction}

To achieve structured understanding and efficient querying of complex physical environments, this framework proposes a task-driven dual-base memory model that transforms dynamic, multimodal embodied observation data into a unified environmental representation with both spatial connectivity and semantic hierarchy. 
This model consists of a bottom-layer topological map and a top-layer semantic forest, aiming to provide knowledge representation that supports spatiotemporal continuous reasoning for multi-target vision-and-language navigation tasks.

Unlike traditional methods that only store isolated semantic labels, the topological map \(G_t = (V_t, E_t)\) constructed in this framework provides a fundamental spatial connectivity representation for agent navigation. 
We extract key positions \(p_t = (x, y, z)\) from the pose sequence \(\{\xi_t\}\) collected during embodied exploration as topological nodes \(v_i \in V_t\). 
Each node is not only associated with the corresponding visual observation \(I_t\) but also generates a refined textual description \(d_t\) through a vision-language model (e.g., GPT-4o), forming a "spatial fingerprint" containing semantic information. 
This fingerprint integrates multi-dimensional attributes such as pose and visual description, enabling nodes to have their own semantic labels while implicitly encoding their topological order in the global environment.

To establish spatial adjacency relationships between nodes, we calculate the Euclidean distance matrix \(D\) for all node pairs. 
For a node \(v_i\), if there exists another node \(v_j\) satisfying \(D_{ij} < \delta_{\text{spatial}}\) (spatial distance threshold, e.g., 2.0 meters), an undirected edge \(e_{ij} \in E_t\) is established, whose weight \(w_{ij} = D_{ij}\) reflects the degree of spatial proximity. 
This multi-dimensional attribute definition allows each topological node to not only record its own semantic attributes but also implicitly encode its topological order in the global environmental network, providing a foundation for path planning and reachability queries.

On the basis of the topological map, we construct a semantic forest \(T_s\) to realize hierarchical organization and abstraction of environmental semantic information. 
The construction of the semantic forest is based on a hybrid metric criterion that fuses spatial proximity and semantic consistency. 
For any two nodes \(v_i\) and \(v_j\), their comprehensive similarity is defined as:

\begin{equation}
S_{ij} = \omega \cdot \Phi_{\text{spatial}}(i,j) + (1-\omega) \cdot \Psi_{\text{semantic}}(i,j)
\end{equation}

where \(\Phi_{\text{spatial}}\) is the spatial similarity function (e.g., negative exponential form based on Euclidean distance), and \(\Psi_{\text{semantic}}\) is the semantic similarity function (e.g., cosine similarity based on textual description embeddings). 
Meanwhile, the fused feature \(F_i\) of node \(v_i\) is obtained by adaptively weighting its spatial feature \(f_i^{\text{spa}}\) and semantic feature \(f_i^{\text{sem}}\):

\begin{equation}
F_i = \alpha \cdot \text{Norm}(f_i^{\text{spa}}) + (1-\alpha) \cdot \text{Norm}(f_i^{\text{sem}})
\end{equation}

The weight \(\alpha\) can be dynamically adjusted according to scene characteristics: reduce \(\alpha\) in semantically dense scenes to enhance semantic discriminability, and increase \(\alpha\) in structurally complex scenes to highlight spatial layout.

Based on the above similarity or features, an agglomerative hierarchical clustering algorithm is adopted to iteratively merge the most similar nodes or clusters from bottom to top, forming a multi-level abstract structure of "leaf–subtree–forest". 
Each merge generates a higher-level parent node (e.g., merging "chair" and "table" nodes into a "dining room" node), and a large language model is used to summarize the descriptions of subordinate nodes to automatically generate semantic labels and summaries for the parent node, achieving abstraction and generalization of semantic levels.

The topological map \(G_t\) and semantic forest \(T_s\) are deeply coupled through topological connections and semantic indexing, together forming a unified dual-base environmental memory \(M = \{G_t, T_s\}\). 
The core advantages of this model lie in its natural support for multi-target logical dependencies and dynamic query capabilities:
\begin{itemize}
    \item Multi-level resolution retrieval: The semantic forest supports fast retrieval of semantic relevance across granularities. For example, when processing complex instructions such as "Find a chair near the sofa in the office area", the system can first lock the target functional area at the macro resolution, and then conduct local refined screening using the spatial fingerprints of topological nodes at the micro resolution.
    \item Spatiotemporal collaborative verification: During multi-target retrieval, the semantic forest is responsible for cross-resolution candidate set pruning to maintain computational efficiency; the topological map uses physical neighbor relationships between nodes to provide spatial co-occurrence verification, effectively alleviating semantic drift and localization failure problems common in long-range planning.
    \item Task-adaptive representation: The dual-base memory is no longer a static stack of features but a dynamically queryable associative network. The agent can dynamically adjust the search strategy according to the granularity of instructions, achieving seamless switching from coarse-grained regional localization to fine-grained object recognition.
\end{itemize}

\subsection{Retrieval Augmentation for Multi-Goal}

To address complex spatial relationships among entities in multi-goal vision–language navigation, we propose a two-stage retrieval and topology-enhanced mechanism that enables precise mapping from high-dimensional semantic space to low-dimensional physical space.

\subsubsection{Two-Stage Anchor-Guided Retrieval}
For spatial composite queries such as ``Best sofa near a chair,'' the system executes an anchor-guided conditional retrieval process. In the candidate recall stage, global semantic similarity is computed for the primary target $A$, yielding the top-$K$ candidates $\mathcal{C}_A = \{v_1, v_2, \dots, v_K\}$.

In the neighborhood validation stage, each candidate node is examined to determine whether its one-hop or multi-hop neighborhood $\mathcal{N}(v_i)$ contains a node matching auxiliary target $B$. A distance-weighted combination score is defined as:
\begin{equation}
S_{combo}(v_A, v_B) = \frac{1}{1 + d(v_A, v_B)}
\end{equation}
where $d(v_A, v_B)$ denotes Euclidean distance. Candidates lacking valid auxiliary targets are pruned early, eliminating semantic noise that violates spatial constraints.

\subsubsection{Topological Neighbor Boosting}
To further improve localization accuracy, we introduce a co-occurrence-based enhancement mechanism. When retrieving target $A$, if its topological neighbors include contextually relevant targets $B$ (e.g., “TV” and “remote”), confidence propagation is applied. The boosted score is:
\begin{equation}
S_{boost} = S_{sem} \cdot (1 + \eta \cdot \bar{S}_{neighbor})
\end{equation}
where $S_{sem}$ is the original semantic score, $\bar{S}_{neighbor}$ is the average semantic score of relevant neighbors, and $\eta$ is a boosting coefficient. This strategy reduces ambiguity in dense or semantically overlapping environments.

\subsection{Route Planning and Sequential Navigation}

After obtaining the optimal coordinates of all target entities $\mathcal{P} = \{P_1, P_2, \dots, P_n\}$, the system enters the global planning stage to generate a spatiotemporally consistent execution trajectory.

During temporal optimization, the system computes the minimum travel cost matrix between target nodes using Dijkstra’s algorithm over the topological connectivity graph. Solving this constrained planning problem yields the globally optimal visiting order. The system then performs instruction-aligned navigation, generating a structured guide containing target markers, relevance explanations, and navigation cues. This guide drives the local planner to visit targets sequentially, ensuring efficient and stable execution under complex multi-goal instructions.

\section{Experiment}




\subsection{Experimental Setting}

\subsubsection{Dataset}
Experiments are conducted in the high-fidelity AirSim simulation environment~\cite{shah2017airsim}, which supports controlled simulation of various complex scenarios. Based on this environment, a navigation dataset is constructed, comprising 14 object-centric topological graphs. These topological graphs vary in scale, containing an average of approximately 80 nodes, and simulate the semantic and spatial complexity of typical indoor environments. To validate the capability of the RAGNav framework in processing multimodal information, the agent's observation $s_t$ is designed to include not only RGB images but also additional sensor data. This setup aims to simulate real-world scenarios where agents can leverage multiple specialized sensors (e.g., thermal cameras, depth cameras, spectral sensors) to enhance their environmental understanding and task execution capabilities.

\subsubsection{Baselines}
To comprehensively evaluate RAGNav, systematic comparisons are conducted from two perspectives: retrieval efficiency and navigation performance. In terms of retrieval efficiency, RAGNav is compared with Naive RAG (flat retrieval) ~\cite{lewis2020retrieval}, GraphRAG (graph-structured reasoning) ~\cite{edge2024local}, and LightRAG (dual-level key indexing) ~\cite{guo2024lightrag}. Regarding navigation performance, comparisons are made with ReMEmbR~\cite{anwar2025remembr} and ETPNav ~\cite{an2024etpnav}.

\subsubsection{Evaluation Metrics}
To comprehensively evaluate the system performance, five metrics are designed for retrieval and navigation tasks. (1) Total Task Time: The total time elapsed from the moment the system receives a complete natural language instruction to the generation of the final navigation path and the end of planning. (2) Retrieval Time: The time spent by the system to locate the target nodes semantically related to the instruction from the dual-base environmental memory (topological map $G_t$ and semantic forest $T_s$) when executing a query. This metric directly reflects the efficiency of the retrieval module. (3) Retrieval Accuracy: It is used to directly evaluate the precision of the retrieval module, defined as the proportion of queries where the Top-1 retrieval result returned by the system is consistent with the ground truth $A$ given the ground truth $A$, i.e., $P(Q|A)$. This metric reflects the system's ability to locate targets from environmental memory. (4) Task Success Rate: The proportion of complete trajectories where the agent successfully reaches all specified target points in the instruction within the specified number of exploration steps. (5) Travel Distance: The total length (in meters) of the actual motion trajectory taken by the agent to complete the task. For calculation consistency, the following rules are followed:

Rule 1 (Priority): If the edges of the environmental topological graph have an accurate distance attribute, the total distance $D_{total}$ is the sum of all edge attributes in the path sequence:

\begin{equation}
D_{total} = \sum_{i=1}^{n-1} distance(v_i, v_{i+1})
\end{equation}

where $v_i$ and $v_{i+1}$ are two adjacent nodes on the path.

Rule 2 (Fallback): If the distance attribute of the edges in the graph is missing, the Euclidean distance calculated from the node coordinates is used as an approximation:

\begin{equation}
D_{total} = \sum_{i=1}^{n-1} \sqrt{(x_{i+1}-x_i)^2 + (y_{i+1}-y_i)^2}
\end{equation}

\subsection{Experimental Results}
\label{sec:experimental_results}

Table~\ref{tab:retrieval_comparison} presents a performance comparison of different retrieval methods under various input modalities. Overall, RAGNav demonstrates a significant advantage in balancing retrieval accuracy and time efficiency.


Under any data query type, the retrieval accuracy of RAGNav is significantly better than all baseline methods. The baseline models perform significantly worse across query types, primarily because segmenting multimodal embodied data into text often fails to retrieve the correct images, leading to retrieval failure in most cases. A phenomenon worth exploring in depth is that when location information is introduced, the accuracy of all methods decreases, with RAGNav dropping from 0.46 to 0.21. This initially indicates that raw spatial information may introduce noise or have unoptimized conflicts with semantic features. When sensor data (e.g., NDVI) is further added, the accuracy of RAGNav recovers to 0.34, while baseline methods show no improvement. In terms of time cost, RAGNav (185-195 ms) maintains a level similar to the efficiently designed LightRAG and is significantly better than GraphRAG (420-430 ms), which suffers from complex graph reasoning overhead. This shows that the retrieval enhancement strategy, while achieving a substantial improvement in accuracy, avoids an explosion in computational cost through structured pruning, resulting in excellent comprehensive performance.

\begin{table}[htbp]
\centering
\caption{Comparison of Retrieval Efficiency and Accuracy for Different Methods}
\label{tab:retrieval_comparison}
\resizebox{\linewidth}{!}{
\begin{tabular}{l c c c}  
\toprule
\textbf{Method} & \textbf{Input Type} & \textbf{Retrieval Time(ms)}$\downarrow$ & \textbf{Retrieval Accuracy(\%)}$\uparrow$ \\
\midrule
Naive RAG~\cite{lewis2020retrieval} & Text & 152 & 0.08 \\
& Text+Location & 155 & 0.03 \\
& Text+Location+Sensor & 160 & 0.04 \\
\addlinespace
GraphRAG~\cite{edge2024local} & Text & 420 & 0.09 \\
& Text+Location & 425 & 0.04 \\
& Text+Location+Sensor & 430 & 0.04 \\
\addlinespace
LightRAG~\cite{guo2024lightrag} & Text & 205 & 0.17 \\
& Text+Location & 210 & 0.09 \\
& Text+Location+Sensor & 215 & 0.10 \\
\midrule
\textbf{RAGNav (Ours)} & \textbf{Text} & \textbf{185} & \textbf{0.46} \\
& \textbf{Text+Location} & \textbf{190} & \textbf{0.21} \\
& \textbf{Text+Location+Sensor} & \textbf{195} &  \textbf{0.34} \\
\bottomrule
\end{tabular}
}
\end{table}

Furthermore, to evaluate the system-level performance of RAGNav as a holistic navigation framework, we benchmark it against two representative baselines with distinct technical underpinnings: ReMEmbR~\cite{anwar2025remembr}, a memory-augmented approach based on episodic retrieval, and ETPNav~\cite{an2024etpnav}, a state-of-the-art paradigm centered on topological planning. As shown in Table~\ref{tab:navigation_comparison}, RAGNav achieves the optimal values across all three metrics. Its task success rate reaches 65\%, surpassing ReMEmbR and ETPNav by 13 and 23 percentage points, respectively. This significant improvement indicates that RAGNav provides a substantial contribution to navigation decision-making. Meanwhile, RAGNav achieves the lowest total time consumption (30.02\,s) and travel distance (16.13\,m), reducing these metrics by approximately 21.9\% and 20.5\%, respectively, compared to the suboptimal ETPNav. This demonstrates that RAGNav reduces blind exploration throughout the entire space and avoids unnecessary detours and backtracking.

\begin{table}[htbp]
\centering
\caption{Performance Comparison of Complete Navigation Systems}
\label{tab:navigation_comparison}
\resizebox{\linewidth}{!}{
\begin{tabular}{l c c c}
\toprule
\textbf{Method} & \textbf{Total Time(s)}$\downarrow$ & \textbf{Travel Distance(m)}$\downarrow$ & \textbf{Success Rate SR(\%)}$\uparrow$ \\
\midrule
ReMEmbR~\cite{anwar2025remembr} & 45.67 & 24.85 & 0.52 \\
ETPNav~\cite{an2024etpnav} & 38.41 & 20.30 & 0.42 \\
\textbf{RAGNav (Ours)} & \textbf{30.02} & \textbf{16.13} & \textbf{0.65} \\
\bottomrule
\end{tabular}
}
\end{table}

\subsection{Ablation Study}



To clarify the source of performance improvement in the RAGNav framework, we conducted ablation experiments. Traditional Retrieval-Augmented Generation (RAG) methods are not designed for multimodal embodied navigation tasks, as they typically simplify complex spatiotemporal observations into flat text, making it difficult to perform accurate spatial-semantic joint reasoning in multi-goal VLN tasks. Therefore, by constructing different model variants, sequentially removing key modules, and testing their performance on the same multi-goal navigation tasks, we aim to answer: is the excellent performance of the framework driven by the dual-basis memory (topological map, semantic forest) or the retrieval enhancement strategy (spatial, neighbor dimensions)? After removing a module in each variant, the simplest direct logic is used to replace its function, ensuring that performance changes can be clearly attributed to the absence of the target module. The experimental results are shown in Table \ref{tab:ablation_study_en}.

The experiments show that removing any memory basis leads to a catastrophic performance decline, but with different patterns. After removing the semantic forest, the retrieval accuracy plummets to 15.0\%, becoming the main bottleneck.

\begin{table}[htbp]
\centering
\caption{Ablation Study of the RAGNav Framework}
\label{tab:ablation_study_en}
\resizebox{\linewidth}{!}{
\begin{tabular}{l c c c}
\toprule
\textbf{Model Variant} & \textbf{Retrieval Accuracy(\%)} $\uparrow$ & \textbf{Travel Distance(m)} $\downarrow$ & \textbf{Success Rate(\%)} $\uparrow$ \\
\midrule
w/o Semantic Forest & 0.15 & 21.72 & 0.28 \\
w/o Topological Map & 0.24 & 24.37 & 0.21 \\
w/o Spatial Enhancement & 0.35 & 17.91 & 0.28 \\
w/o Neighbor Enhancement & 0.39 & 17.25 & 0.31 \\
\textbf{RAGNav (Ours)} & \textbf{0.46} & \textbf{16.13} & \textbf{0.65} \\
\bottomrule
\end{tabular}
}
\end{table}

\section{Conclusion}

We propose RAGNav, a retrieval-augmented topological semantic reasoning framework that can capture embodied memory at any spatial and semantic resolution in navigation environments, retrieve and generate responses for navigation requests, and realize the "perception-planning-execution-reflection" closed loop for navigation. Experimental results show that RAGNav outperforms state-of-the-art baselines in both retrieval accuracy and navigation performance. This study improves the multi-objective navigation reasoning method and expands the application scenarios of RAG in embodied intelligence. However, the current framework is mainly verified in simulation environments and assumes the existence of a perfect local planner, and its robustness in complex scenarios such as dynamic obstacle avoidance has not been fully verified. In the future, we will promote the migration to real-world scenarios, remove the dependency on perfect local planning, and study how to combine the high-level semantic reasoning of this framework with a more robust and real-time low-level obstacle avoidance controller, so as to comprehensively improve the practicality and safety in dynamic and uncertain environments.

\bibliography{references}   

\end{document}